\documentclass{article}

\usepackage[preprint]{neurips_2024}

\usepackage[utf8]{inputenc} % allow utf-8 input
\usepackage[T1]{fontenc}    % use 8-bit T1 fonts
\usepackage{hyperref}       % hyperlinks
\usepackage{url}            % simple URL typesetting
\usepackage{booktabs}       % professional-quality tables
\usepackage{amsfonts}       % blackboard math symbols
\usepackage{nicefrac}       % compact symbols for 1/2, etc.
\usepackage{microtype}      % microtypography
\usepackage{xcolor}         % colors
\usepackage{graphicx}
\usepackage{amsmath}
\usepackage{multirow} 
\usepackage{booktabs}
\usepackage{floatrow}
\title{Redefining Automotive Radar Imaging: A Domain-Informed 1D Deep Learning Approach for High-Resolution and Efficient Performance}

\setcitestyle{round, numbers}

\author{%
  Ruxin~Zheng
    \\
  The University of Alabama\\
  Tuscaloosa, AL 35487 \\
  \And
  Shunqiao~Sun \\
  The University of Alabama\\
  Tuscaloosa, AL 35487 \\ 
  \AND
   Holger~Caesar \\
  Delft University of Technology \\
  Delft, Netherlands \\
  \And
  Honglei~Chen \\
  Mathworks, Inc \\
  Natick, MA 01760 \\
  \And
  Jian~Li \\
  University of Florida \\
  Gainesville, FL 32611 \\
}

\begin{document}

\maketitle

\begin{abstract}
Millimeter-wave (mmWave) radars are indispensable for perception tasks of autonomous vehicles, thanks to their resilience in challenging weather conditions. Yet, their deployment is often limited by insufficient spatial resolution for precise semantic scene interpretation. Classical super-resolution techniques adapted from optical imaging inadequately address the distinct characteristics of radar signal data. In response, our study redefines radar imaging super-resolution as a one-dimensional (1D) signal super-resolution spectra estimation problem by harnessing the radar signal processing domain knowledge, introducing innovative data normalization and a domain-informed signal-to-noise ratio (SNR)-guided loss function.  Our tailored deep learning network for automotive radar imaging exhibits remarkable scalability, parameter efficiency and fast inference speed, alongside enhanced performance in terms of radar imaging quality and resolution. Extensive testing confirms that our SR-SPECNet sets a new benchmark in producing high-resolution radar range-azimuth images, outperforming existing methods across varied antenna configurations and dataset sizes. Source code and new radar dataset will be made publicly available online.
\end{abstract} 
\section{Introduction}
\label{sec:intro}
Radar technology, particularly in the form of millimeter wave radars, has become a cornerstone for advanced driver assistance systems (ADAS) and autonomous vehicles, surpassing the capabilities of traditional RGB cameras and LiDAR in challenging weather and low visibility conditions \citep{Patole_SPM_2017,engels2017advances,SUN_SPM_Feature_Article_2020,waldschmidt2021automotive,Peng_book_2022,zheng2023deep}. Its adoption is largely driven by the robust, cost-effective, and reliable sensing solutions it offers, operational under virtually all environmental scenarios. Frequency-modulated continuous-wave (FMCW) signals within the millimeter-wave band are primarily utilized in these radar systems, chosen for their cost-efficient operation and potential for high-resolution sensing. This technological choice is pivotal for a broad spectrum of autonomous driving functionalities, including free space detection, \(360^\circ\) surrounding sensing, object detection and classification, and simultaneous localization and mapping (SLAM)~\citep{sun20214d,duggal2020doppler,engels2021automotive}.

Historically, automotive radar technology, dating back to the late 1990s and early 2000s, was developed with a focus on supporting ADAS functions like adaptive cruise control (ACC) \citep{waldschmidt2021automotive}. However, these radar systems primarily measure speed and range, offering limited azimuth angular resolution. To achieve Level 4 and Level 5 fully autonomous driving capabilities, a demand for high-resolution four-dimensional (4D) sensing has emerged~\citep{sun20214d}. Such advanced sensing is essential not only for speed and range determination but also for accurately estimating targets' azimuth and elevation with high resolution.

However, these radar systems primarily measure speed and range, offering limited azimuth angular resolution. The challenge of enhancing angular resolution has led to the extensive use of multiple-input multiple-output (MIMO) radar technology. MIMO radars synthesize a large virtual array aperture, significantly improving angular resolution with a manageable number of transmit and receive antennas \citep{Jian_07,Jian_MIMO_book,SUN_SPM_Feature_Article_2020,Guerci_MIMO_book_2018}.

\begin{figure}%[ht]
\centering
\includegraphics[width= 1.0\linewidth]{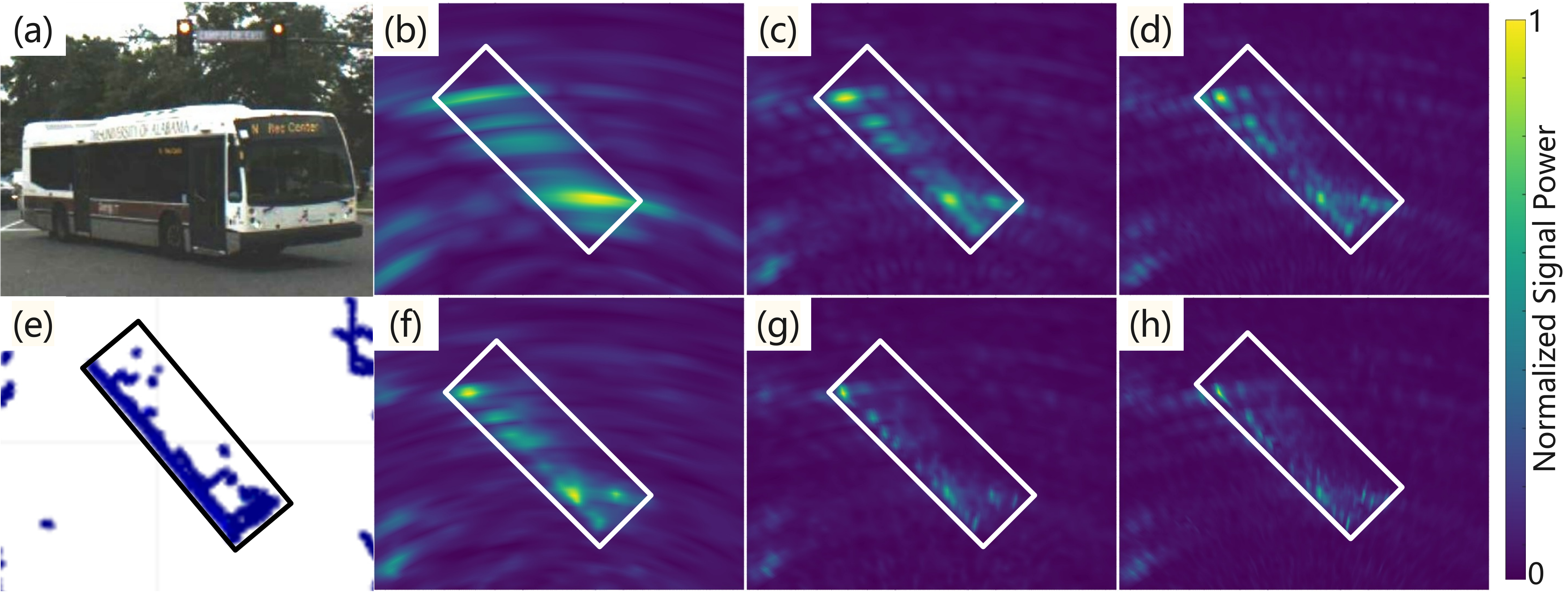}
\caption{Impact of antenna aperture and super-resolution algorithms on RA heatmap quality: (a) shows an RGB bus image. RA heatmaps using FFT for (b) 10, (c) 40, and (d) 86 antennas contrast with (e) LiDAR BEV. Heatmaps with IAA for (f) 10, (g) 40, and (h) 86 antennas highlight the improved clarity from both antenna count and super-resolution algorithm.}
\label{fig1}
\end{figure}

Further, signal processing techniques have been explored to further improve the angular resolution beyond what is achievable through digital beamforming technique that is implemented via fast Fourier transform (FFT). Super-resolution direction of arrival (DOA) estimation algorithms, such as compressive sensing (CS) \citep{donoho2006compressed,candes2007sparsity,Candes_Super_Resolution_2014} and the iterative adaptive approach (IAA) \citep{Yardibi_IAA_2010,Roberts_IAA_2010}, represent significant strides in this direction. Yet, their computational demand presents a formidable barrier to real-time implementation, especially in the dynamic context of automotive scenarios. Figure \ref{fig1} illustrates how antenna aperture and super-resolution algorithms influence the quality of range-azimuth (RA) heatmaps. The high-resolution RA heatmaps contain rich information of the objects, including their shapes, facilitating object detection and classification through deep neural networks \citep{zheng2023deep}.

The adoption of deep learning (DL) techniques for radar image enhancement has yielded significant advances within the realm of image super-resolution, as demonstrated in computer vision research \citep{8902510,geiss2020radar,li2023azimuth}. Applying these methods to enhance azimuth resolution in RA heatmaps presents a significant opportunity for substantial improvement. Nonetheless, few studies have focused on generating super-resolution RA heatmaps using raw radar signals by exploiting radar domain knowledge. Approaches that consider generating radar super-resolution RA heatmaps as straightforward image-to-image or volume-to-volume tasks often overlook the critical domain knowledge of radar signal processing. This oversight can lead to data-intensive solutions, rely on excessively large networks, or fail to deliver optimal performance and scalability which are key concerns in automotive applications where rapid inference and compact model size are essential for on-chip implementation. 

Research in the domain of super-resolution RA heatmap generation for automotive radar remains limited, with the majority of studies relying on FFT-generated ground truths from larger antenna arrays. To our best knowledge, no existing methods leverage RA heatmaps produced through super-resolution algorithms as ground truths. Moreover, these methods typically focus on smaller antenna arrays and do not explore the potential of varied training data sizes. This paper aims to close these gaps by introducing the Super-Resolution Angular Spectra Estimation Network (SR-SPECNet). Designed with radar signal processing expertise, SR-SPECNet advances super-resolution angular spectra generation by transforming RA heatmap enhancement into a manageable 1D azimuth super-resolution challenge. This transformation is supported by our novel data normalization approach and an signal-to-noise ratio (SNR)-guided loss function. SR-SPECNet is thoroughly evaluated across varied antenna apertures and training dataset sizes, a first in this research area, using a dedicated real-world dataset. Our  experimental analysis confirms that SR-SPECNet achieves exceptional parameter efficiency, superior performance in imaging quality, and outstanding scalability. It consistently surpasses established benchmarks, showcasing its capability to adapt to various antenna configurations and dataset sizes.

The key contributions of our work include:
\begin{itemize}
\item We introduced SR-SPECNet, a network designed for efficiency and effectiveness, capable of using single-snapshot measurement to robustly produce high-resolution automotive RA imaging typically obtained through IAA, but without IAA's computational expense.

\item We adopted radar signal processing domain knowledge to guide the neural network design by translating the RA imaging as a 1D spectra estimation problem, introducing a novel real radar data normalization method and a SNR-guided loss function.

\item SR-SPECNet is the first network proven to robustly create high-resolution RA imaging with fast inference time from real automotive radar data featuring dynamic objects, demonstrating SR-SPECNet's scalability, efficiency, and robust performance.
\end{itemize}

\section{Related Work}
\label{sec:relate}

The quest for enhanced radar imaging  has predominantly focused on improving the azimuth resolution with limited number of antenna elements, given that range resolution can be augmented by increasing the  bandwidth. In the automotive radar domain, digital beamforming (DBF) has emerged as the predominant DOA estimation algorithm, favored for its computational efficiency and robustness. This technique, typically implemented via FFT, however, faces limitations in angular resolution due to the Rayleigh criterion and is characterized by relatively high sidelobes \citep{SUN_SPM_Feature_Article_2020,Richards}. Automotive radars, operating within highly dynamic environments, often have access to only a limited number of snapshots, sometimes as few as a single snapshot. 
This scenario renders super-resolution methods, such as Capon beamforming, MUSIC \citep{schmidt1982signal}, and ESPRIT \citep{Kailath_ESPRIT_1989}, which require multiple snapshots for  covariance matrix estimation, less viable.

Compressive sensing (CS) techniques, which leverage the sparsity of target distributions in the angular domain, are highly effective in super-resolution, especially in snapshot-constrained settings \citep{donoho2006compressed,Candes_Super_Resolution_2014}. Despite their potential, CS methods demand a dictionary matrix with low mutual coherence, which can be limiting. Alternatively, the Iterative Adaptive Approach (IAA) offers robust Direction of Arrival (DOA) estimation with limited snapshots, utilizing a nonparametric, iterative process \citep{Yardibi_IAA_2010,Roberts_IAA_2010}. IAA is advantageous for high-resolution radar imaging, surpassing subspace methods like MUSIC and ESPRIT and CS-based methods that falter under snapshot constraints or produce sparse results. However, IAA is computationally intensive, requiring large-scale matrix inversions at each step. 
Although fast and super-fast IAA variants \citep{Glentis_Fast_IAA_2011,Xue_Fast_IAA_2011, Glentis_Superfast_IAA_2012} aim to reduce these demands by replacing matrix inversions with factorization, their benefits are marginal for small arrays and more pronounced for larger arrays, though computational challenges persist.

Recently, deep learning techniques have been applied to address the intricacies of azimuth super-resolution in RA maps. 
An adversarial network was tailored for super-resolution in micro-Doppler imagery \citep{8902510}, showcasing the potential of generative adversarial networks (GANs) in radar image enhancement. A U-Net architecture was employed for the super-resolution of weather radar maps \citep{geiss2020radar}, demonstrating the adaptability of deep convolutional networks to various radar data modalities. Notably, \citep{li2023azimuth} ventured into extrapolating received antenna signals through a compact network, followed by the application of a 3D U-Net on the range-Doppler-azimuth data cube, facilitating the generation of super-resolution RA heatmaps. 
However, the existing body of work primarily leverages 2D or 3D network architectures, predicated on the assumption that the problem space necessitates multi-dimensional data processing to achieve enhanced resolution. This perspective, while valid, overlooks the potential efficiencies and novel insights that can be garnered from reinterpreting the challenge through a one-dimensional lens. To our best knowledge, no prior work has endeavored to address radar azimuth super-resolution within RA heatmaps using a 1D approach. 
In this paper, we close this gap by designing an efficient and effective deep neural network to achieving super-resolution radar imaging, through leveraging radar signal processing domain knowledge.

\section{Radar Datasets}
\label{sec:data}

\begin{table*}[ht]
\centering
\resizebox{\textwidth}{!}{%
 
\begin{tabular}{lllll}
 \toprule
{\bf Dataset} & {\bf \# of Frames} & {\bf Data Type} & {\bf Resolution} & {\bf Radar/Technology} \\ \midrule
nuScenes \citep{nuScenes}  & $40,000$ & Sparse PC  & Low & Continental ARS408 \\ \midrule
Oxford Radar  \citep{Oxford_radar_robotcar} & $240,000$  & RA    & High & Navtech Spinning Radar\\ \midrule
RADIATE \citep{RADIATE} & $44,000$ & RA  & High  & Navtech Spinning Radar\\ \midrule
 CRUW \citep{CRUW} & $396,241$ & RA    & Low & TI AWR1843\\ \midrule
 Zendar \citep{Zendar} & $94,460$ &  ADC,RD,PC   & High  & SAR\\ \midrule
CARRADA \citep{CARRADA} & $12,666$ & RA,RD,RAD  & Low & TI AWR1843\\ \midrule
 RadarScenes \citep{RadarScenes} & $975$ &  Dense PC  & High & 77GHz Middle-Range Radar\\ \midrule
RADIal\citep{valeo} & $25,000$ & ADC,RD,PC    &  High & Valeo Middle Range DDM\\ \midrule
View-of-Delft\citep{apalffy2022} & $8,693$ & PC+Doppler  &  High & ZF FRGen21 Radar\\ \midrule
% Zendar \citep{Zendar} & $94,460$ &  ADC,RD,PC   & High  & SAR\\ \midrule
K-Radar \citep{paek2022kradar} & $35,000$ &  4D Tensor  &  High  & KAIST-Radar\\ \midrule
Radatron \citep{radatron} & $152,000$ &  4D Tensor  &  High  & TI Cascade Imaging Radar\\ \midrule
Ours  & $17,000$ &  ADC  &  High  & TI Cascade Imaging Radar\\ \bottomrule
\end{tabular}}
\caption{Overview of publicly available radar data sets. Data Type: Raw ADC data (ADC), Range-Doppler map (RD), Range-Azimuth map (RA), point clouds (PC).}
\label{table_1}
\end{table*}

Radar datasets for autonomous driving such as nuScenes \citep{nuScenes}, Oxford Radar RobotCar \citep{Oxford_radar_robotcar}, RADIATE \citep{RADIATE}, and others, are summarized in Table \ref{table_1}. 
Technologies like spinning radar, utilized in the RADIATE and Oxford Radar RobotCar datasets, provide high-resolution 360-degree field-of-view (FOV) imagery, albeit at limited frame rates, which can introduce motion blur challenges. Dataset such as CARRADA employs single-chip Texas Instruments (TI) radar systems, offering modest angular resolutions exceeding $10^\circ$. 
The Zendar dataset, leveraging synthetic aperture radar (SAR) technology, excels in imaging static targets by integrating measurements from different vehicle positions. 
The View-of-Delft dataset takes advantage of the ZF FRGen21 radar's long-range and high-resolution imaging capabilities, offering point cloud data with object annotations confined to a 50 meter range.

\subsection{Our Dataset}
\begin{figure}%[h]
\centering
\includegraphics[width=1.0\linewidth]{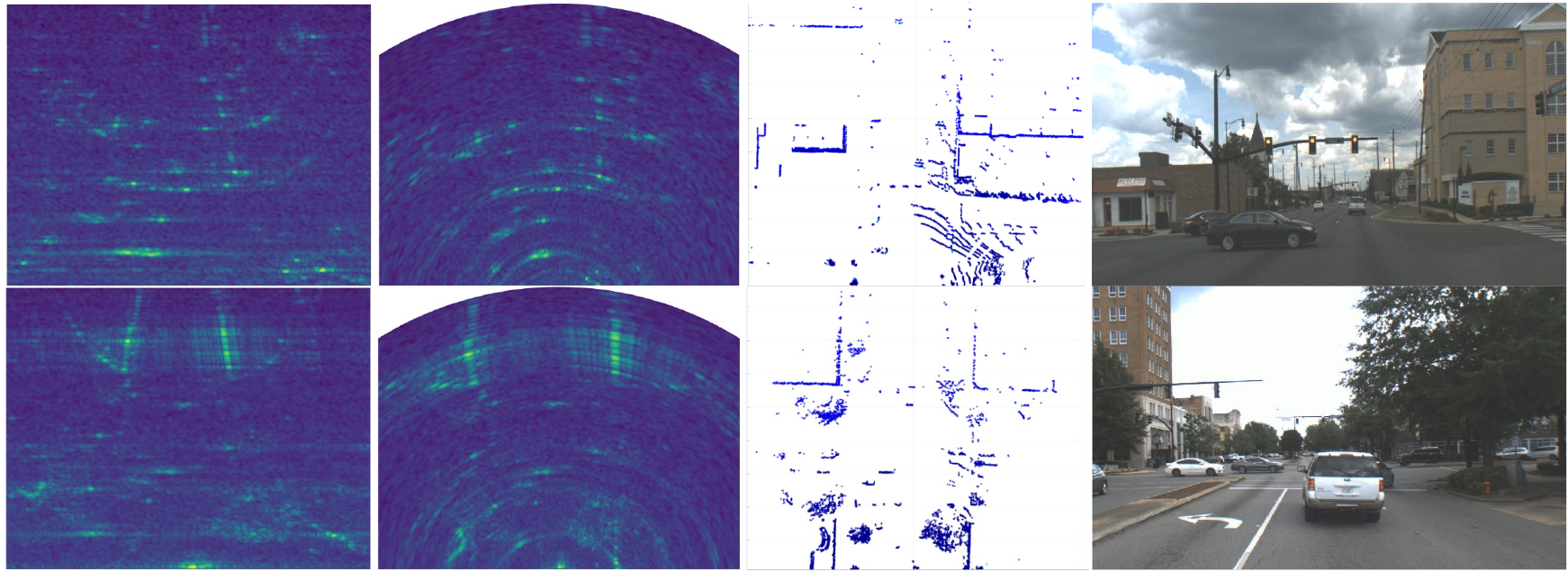}

\caption{\label{fig_radar_BEV_example} 
Left to right columns: RA maps in polar coordinates, RA maps in Cartesian coordinates, LiDAR point clouds in bird's-eye view, and camera image.
}
\end{figure}

Our approach demands detailed radar configuration parameters and intensive raw analog-to-digital converter (ADC) data processing to integrate super-resolution algorithms effectively and assess network performance across various antenna apertures. Hence, we created  our own dataset by driving a Lexus RX450h SUV equipped with multi-modal sensors, including a TI imaging radar, Teledyne FLIR Blackfly S stereo cameras, and a Velodyne Ultra Puck VLP-32C LiDAR sensor, along urban streets, highways and campus roads. The centerpiece of our dataset is the TI cascaded imaging radar system \citep{TI_Cascade}, configured for MIMO operations with an array of 12 transmit (TX) and 16 receive (RX) antennas.
The operational 9 TX and 16 RX antennas were arranged to form a virtual uniform linear array (ULA) of 86 elements, with half-wavelength spacing, rendering an azimuth resolution of roughly 1.2 degrees via FFT. 
Our dataset showcases the exceptional high-resolution capabilities of our radar configuration, as illustrated in Figure \ref{fig_radar_BEV_example}. 

\section{Method}
\label{sec:propose}

We aim to transform raw ADC data into high-resolution RA maps. Unlike recent approaches that derive high-resolution ground truth from RA maps using an expanded antenna array \citep{li2023azimuth}, our method relies on RA maps generated with the same number of antennas but refined using IAA algorithms as our benchmark.

\begin{figure}[]
\centering
\includegraphics[width=1.0\linewidth]{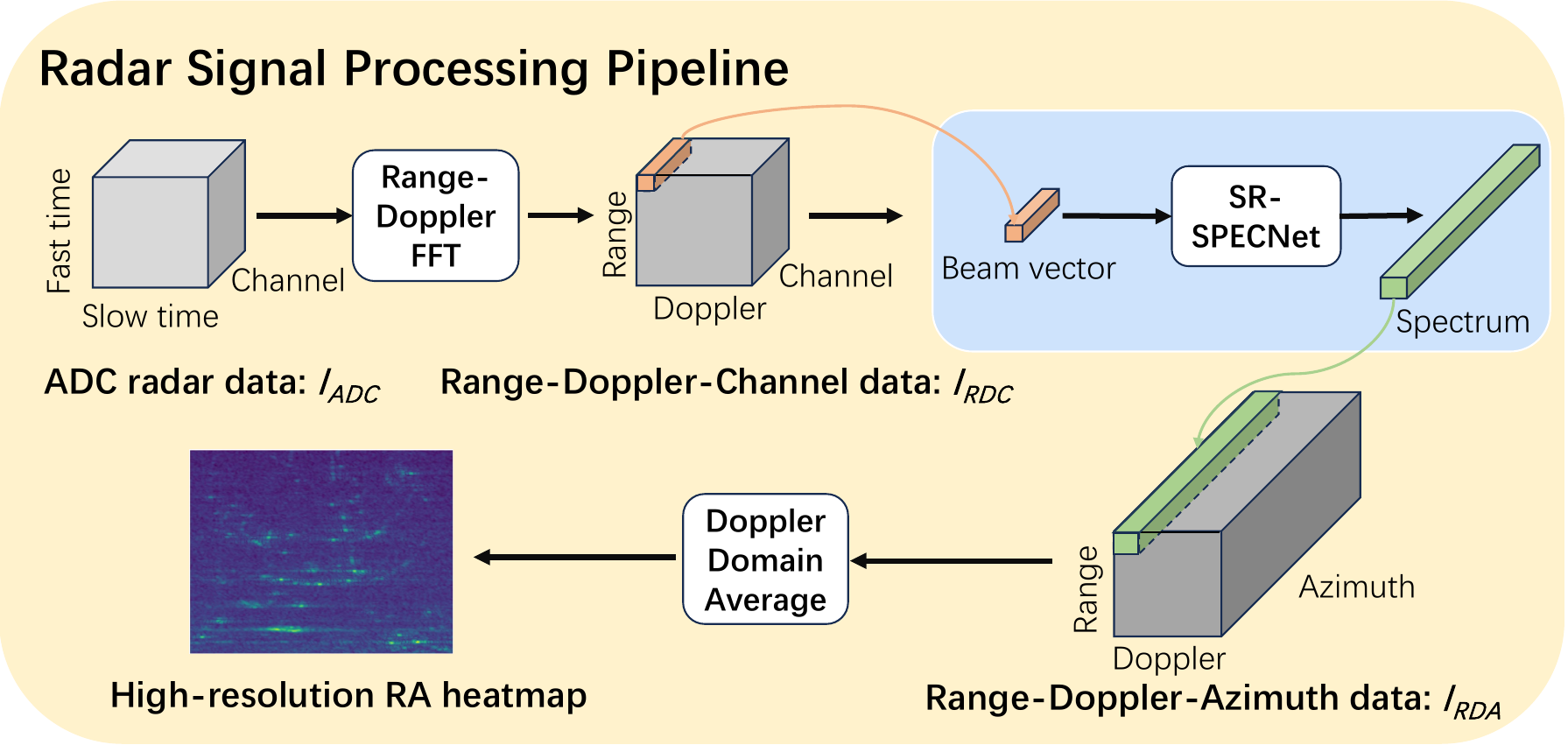}
\caption{\label{pip}Radar signal processing pipeline}
\vspace{-3mm}
\end{figure}

Figure \ref{pip} depicts our processing workflow. The input ADC data, $I_{\rm ADC} \in \mathbb{C}^{N_{\rm fast} \times N_{\rm slow} \times N_{\rm ch}}$, encapsulates three dimensions: $N_{\rm fast}$ for fast time samples, $N_{\rm slow}$ for slow time samples (or chirps), and $N_{\rm ch}$ for channels (or receivers). Through a 2D FFT to $I_{\rm ADC}$ across both fast and slow time dimensions, we obtain range-Doppler-channel data, $I_{\rm RDC} \in \mathbb{C}^{N_{\rm Range} \times N_{\rm Doppler} \times N_{\rm ch}}$. Subsequently, beam vectors $\textbf{y} \in \mathbb{C}^{1 \times 1 \times N_{\rm ch}}$ are extracted from each range-Doppler bin. These vectors are processed by SR-SPECNet to generate a super-resolution spectrum. 
This operation, performed on all beam vectors across all range-Doppler bins, yields the range-Doppler-azimuth data, $I_{\rm RDA} \in \mathbb{R}^{N_{\rm Range} \times N_{\rm Doppler} \times N_{\rm Azimuth}}$. Notably, this procedure is highly parallelizable, treating the dataset as a 2D matrix with a batch size of ${N_{\rm Range} \times N_{\rm Doppler}}$. The final high-resolution RA maps, $M \in \mathbb{R}^{N_{\rm Range} \times N_{\rm Azimuth}}$, are achieved by averaging $I_{\rm RDA}$ over Doppler dimension.

To provide a clearer understanding of the beam vector in the context of automotive radar signal processing, its signal model can be articulated as follows:
\begin{equation}
    \begin{aligned}\label{sig}
    {\bf y} 
         = \textbf{A}(\theta) \textbf{s} +\textbf{n},
    \end{aligned}
\end{equation}
where $\theta$ encapsulates the DOA of targets, $\bf{n}$ signifies a complex $N_{ch} \times 1$ white Gaussian noise vector, and ${\bf A}(\theta) = \left [{\bf a}(\theta_1), {\bf a}(\theta_2), \cdots, {\bf a}(\theta_K) \right]$ represents the $N_{ch} \times K$ array manifold matrix for $K$ targets. The array response vector $\textbf{a}(\theta)$ is given as ${\textbf a}(\theta) = \left [1, e^{\frac{2\pi d_2}{\lambda}\sin{\theta}}, \cdots , e^{\frac{2\pi d_{N_{ch}}}{\lambda}\sin{\theta}} \right ]^T$. In this model, $d_n$ denotes the spacing between the $n$-th element and the reference element, and ${\bf s} = [s_1, s_2, \cdots, s_K]^T$ represents the vector of source strengths.

%\begin{align}

%\end{align}

\subsection{SR-SPECNet for Azimuth Super-Resolution}
CS and IAA stand out as widely recognized super-resolution DOA estimation algorithms tailored for single-snapshot radar data. In contrast to CS, which yields spectra consisting solely of discrete points, the IAA generates continuous spectra that include estimated reflection coefficients. This particular feature renders IAA more apt for the generation of RA maps.

As a super-resolution DOA estimation algorithm working under single-snapshot, IAA generates continuous spectra that include estimated reflection coefficients, rendering IAA more apt for the generation of RA maps.

\subsubsection{Iterative Adaptive Approach (IAA)} 

IAA  is a data-dependent, nonparametric algorithm \citep{Yardibi_IAA_2010}. By discretizing the DOA space into an $L$ point grid,  the array manifold is defined as $\mathbf{A}(\theta) = \left[ \mathbf{a}(\theta_1), \cdots, \mathbf{a}(\theta_L) \right]$ with ${\textbf a}(\theta)$ being the array steering vector. The fictitious  covariance matrix of $\bf{y}$ is represented as ${\bf R}_{f} = {\bf A}(\theta){\bf P}{\bf A}^{H}(\theta)$, where $P$ is a $L \times L$ diagonal matrix with the $l$-th diagonal element being $P_l = |{\hat s}_l|^2$, and ${\hat s}_l$ is the source reflection coefficient corresponding to direction $\theta_l$. IAA iteratively estimates the reflection coefficient $\hat{s}$ and updates the fictitious covariance matrix by minimizing a weighted least-square (WLS) cost function $\lVert {\bf y}-s_{l}{\bf a}(\theta_l) \rVert_{{\bf Q}^{-1}(\theta_l)}^{2}$, where $\lVert {\bf X} \rVert_{{\bf Q}^{-1}(\theta_l)}^{2} \overset{\Delta}{=} {\bf X}^{H}{\bf Q}^{-1}(\theta_l){\bf X}$ and ${\bf Q}(\theta_l) = {\bf R}_f-P_l{\bf a}(\theta_l){\bf a}^{H}(\theta_l)$.

\subsubsection{SR-SPECNet}
SR-SPECNet's primary objective is to transform input beam vectors into super-resolution IAA spectra. The IAA, functioning as an advanced beamforming algorithm, iteratively refines a reconstructed covariance matrix to estimate the spectrum as $\hat{\mathbf{s}}_{\rm IAA} = \mathbf{W}^H\mathbf{y}$, where $\mathbf{W} \in \mathbb{C}^{N_{\rm ch} \times L}$ are the beamforming weights. At each beamforming angle $\theta_L$, the array's response, $\hat{s}_{l} = \mathbf{W}^H(\theta_l)\mathbf{y}$, parallels the output process of a multi-layer perceptron (MLP), underscoring the suitability of using an MLP for this application \citep{naumovski1995neural}.

Designed as a four-layer MLP, SR-SPECNet mirrors the mathematical operations in the IAA algorithm. It processes the input signal
${\bf y }\in {\mathbb C}^{N_{\rm ch}}$, by separating its real and imaginary parts and concatenating them into a real-valued input
${\bf\bar{y} }\in \mathbb{R}^{2*N_{\rm ch}}$. This approach ensures the preservation of crucial phase information, as complex value multiplication inherently involves both the real and imaginary parts.

\subsection{Data Preprocessing} 
Proper data normalization is crucial for training neural networks, especially for regression tasks. 
Different from simulated signals with controlled factors like SNR, target reflection, and target number, real-world signals add unpredictability in SNR and reflections, challenging normalization. SNR varies significantly within a radar frame and from frame to frame. Maintaining a comparable intensity among beam vectors is crucial for constructing accurate RA heatmaps, where factors like SNR, target reflection intensity, and number of targets are precisely controlled. Real-world signals, however, introduce complexities not present in simulated environments: SNR and target reflection intensities vary unpredictably, complicating the task of normalizing inputs and labels. Within a single radar frame, the range of SNR values across beam vectors can be vast, with some vectors lacking targets altogether or presenting very low SNR. Adding to the complexity is the necessity to maintain the relative intensity among beam vectors within each radar frame, a critical aspect for accurately constructing the final RA heatmap.

To overcome the normalization challenges posed by the variability in real-world signals, 
we introduce a frequency domain normalization method designed to produce consistent and interpretable inputs for neural network training. This approach entails determining a normalization factor, $\alpha$, for each beam vector, calculated as the maximum absolute value of the frequency spectra, obtained by multiplying $\mathbf{A}^H$ to the beam vector, equivalent to an FFT operation, and then divided by the total number of elements, $N_{\rm ch}$:
%, within the signal:
\vspace{-3mm}
\begin{align}
    \alpha = \max\left(\left|\frac{\mathbf{A}^H \mathbf{y}}{N_{\rm ch}}\right|\right).
\end{align}

Subsequently, the raw signal $\mathbf{y}$ is normalized using $\alpha$ to yield $\mathbf{y}_{\rm norm} = {\mathbf{y}}/{\alpha}$, ensuring that the signal levels are stable across varying SNR conditions. Similarly, the label, represented by the IAA spectra $\hat{\mathbf{s}}_{\rm IAA}$, is normalized to $\mathbf{s}_{\rm norm} = {\hat{\mathbf{s}}_{\rm IAA}}/{\alpha}$. This normalization strategy effectively scales the signal and the IAA spectra so that their values fall within a comparable range, thereby facilitating more effective network training. 
Further, $\alpha$ maintains a relative intensity among all beam vectors within each radar frame, ensuring that the spatial relationships that are critical for accurate RA heatmap synthesis are preserved. 
Moreover, $\alpha$ preserves the relative intensity across all beam vectors in a radar frame, maintaining the spatial relationships essential for accurate synthesis of RA heatmaps. It's important to note that this normalization process is exclusively needed during training and is not required for generating super-resolution RA heatmaps with test data, which stands as a significant advantage. 
This is attributed to the linear relationship between the beam vector and its corresponding spectra, allowing for direct processing without the need for normalization in the testing phase.
 
\subsection{SNR-Guided Loss Function}
The normalization factor $\alpha$, which represents the maximum value in the signal's frequency domain, is directly proportional to the signal's SNR. A higher 
$\alpha$ suggests a higher SNR, positively influencing the quality of the final RA heatmap. We introduce an SNR-guided loss function similar to a weighted mean squared error (MSE), designed to prioritize higher SNR signals during training. The loss function is defined as:
\vspace{-3mm}
\begin{align}
    \mathcal{L}_{\rm SNR} = \alpha \cdot \frac{1}{L} \sum_{i=1}^{L}(s_i - \hat{s}_i)^2, \nonumber
\end{align}
where $L$ is the number of angle grid points of the spectra, $s_i$ and $\hat{s}_i$ are the actual and predicted values at $\theta_i$. This approach ensures that our model is finely tuned to emphasize higher quality signals.

\section{Experiment}
\label{sec:experiment}

We train and evaluate our SR-SPECNet model using our own dataset, which comprises 17,000 frames of raw ADC radar data. To promote data diversity and minimize the redundancy of consecutive frames, we strategically selected every tenth frame from the dataset, yielding 1,700 frames, with the initial 1,400 frames dedicated to training the model, and the subsequent 300 frames reserved for testing. We intentionally structured the training frames into three subsets to simulate real-world data collection scenarios of limited time periods: a `small' dataset with the initial 200 frames (akin to a 200-second data collection period), a `medium' dataset comprising the first 700 frames, and a `large' dataset that includes all 1,400 frames. This segmentation aims to test our model's performance and adaptability under varying lengths of data availability.

\subsection{Benchmarks}\label{bench}
To evaluate SR-SPECNet's effectiveness, we compare it with models designed to enhance spatial resolution. This comparison includes a 2D U-Net \citep{ronneberger2015u}, which transforms low-resolution RA heatmaps into high-resolution equivalents, and the RAD-UNet \citep{li2023azimuth}, referred to as a 3D U-Net, that upgrades low-resolution range-azimuth-Doppler (RAD) data to high-resolution RAD imagery. We exclude pixel-based super-resolution networks like SRGAN, which increase resolution by adding pixels. These models do not meet the specific requirements of radar imaging, where resolution is not directly related to pixel count \citep{li2023azimuth}.

\subsection{Evaluation Metrics}
The RA map is a grayscale image, normalized between 0 to 1, for both generated and ground truth images. To comprehensively evaluate the quality of high-resolution RA heatmaps, we use established image evaluation metrics. PSNR, measured in dB, and SSIM, ranging from 0 to 1, assess image quality where higher values indicate better quality. NMSE also ranges from 0 to 1 and quantifies prediction accuracy by comparing the mean squared error to the variance of actual values, with lower values indicating more accurate predictions. Together, PSNR, SSIM, and NMSE provide a robust framework for assessing image fidelity, error magnitude, and compositional changes affecting perceived quality. 

\subsection{Implementation Details}\label{sec:experiment:imp}
Our radar configuration is characterized by fast-time samples $N_{\rm fast} = 256$, slow-time samples $N_{\rm slow} = 64$, and post-MIMO processing, resulting in beam vectors, each with $86$ elements. 
We truncated range of the radar data cube, $I_{\rm RDC}$, to keep the first $100$ elements along the range axis, resulting in a truncated dataset $I_{\rm RDC}^{\rm trunc} \in \mathbb{C}^{100 \times 64 \times N_{\rm ch}}$. This truncation strategy is informed by the observation that significant target information is concentrated within the first $50$ meters of the collected data.

We embark on an exploratory analysis of the effect of antenna aperture size on network performance. To this end, we selected a 10-element antenna array to represent a smaller aperture and a 40-element array for a larger aperture. The choice of a 40-element array as the larger aperture is strategically made, considering that it provides sufficient resolution, achieving approximately $1^\circ$ angular resolution using the IAA algorithm.

We set the angular grid size to $L = 256$ for frequency domain uniformity. The labels for our 10-element and 40-element antenna arrays stem from their IAA spectra, ensuring our network's performance evaluation remains consistent across varying apertures. SR-SPECNet comprises four fully connected layers, with the first three followed by ReLU activation functions and output sizes of 2048, 1024, and 512, respectively. The final layer's output size matches $L$. As input signals are complex, we concatenate their real and imaginary parts into a real-valued vector, serving as the input to SR-SPECNet. 

We implemented SR-SPECNet and benchmark models using PyTorch, standardizing training with the Adam optimizer at a learning rate of 0.0001 for 500 epochs. Training was accelerated on four Nvidia RTX A6000 GPUs for efficiency. 

\subsection{High-Resolution RA Heatmap}
We study the performance of deep neural networks in generating high-resolution RA heatmaps. In pursuit of this, we evaluated SR-SPECNet and SR-SPECNet+, which were trained with MSE loss and our SNR-guided loss, respectively, against established benchmark models. As delineated in Tables \ref{10} and \ref{40}, our models were tested using 10 and 40 antenna elements across small, medium, and large dataset sizes.

For the 10-element antenna configuration, Table \ref{10} highlights the strength of our methodology. SR-SPECNet+ emerges as the leading model, eclipsing both the 2D and 3D U-Net benchmarks as well as SR-SPECNet across all dataset sizes. Its edge in performance can be attributed to the incorporation of domain knowledge through the use of an SNR-guided loss function, which fine-tunes the training process to emphasize data with higher signal integrity. This advanced loss function allows SR-SPECNet+ to achieve the lowest NMSE and the highest SSIM and PSNR scores, demonstrating its effectiveness even in small dataset scenarios. SR-SPECNet itself outpaces conventional U-Net models, underscoring the value of generating RA heatmap  through a 1D super-resolution lens. The comparative results make it evident that the domain-specific enhancements in SR-SPECNet+ significantly boost its ability to generate high-quality heatmaps, particularly when the number of antenna elements is limited.

\begin{table}[]
\resizebox{\linewidth}{!}{
\begin{tabular}{clccccccccc}
\toprule
\multicolumn{2}{c}{\multirow{2}{*}{Models}} & \multicolumn{3}{c}{small}                           & \multicolumn{3}{c}{medium}                          & \multicolumn{3}{c}{large}                           \\ \cmidrule{3-11} 
\multicolumn{2}{c}{}                        & NMSE$\downarrow$ & SSIM$\uparrow$ & PSNR$\uparrow$  & NMSE$\downarrow$ & SSIM$\uparrow$ & PSNR$\uparrow$  & NMSE$\downarrow$ & SSIM$\uparrow$ & PSNR$\uparrow$  \\  \midrule
\multicolumn{2}{c}{2D U-Net}                & 0.321                           & 0.782                           & 27.780                           & 0.233                           & 0.820                           & 29.182                           & 0.104                           & 0.877                           & 32.736         \\ \midrule
\multicolumn{2}{c}{3D U-Net}                & 0.763                           & 0.841                           & 26.655                           & 0.205                           & 0.894                           & 31.318                           & 0.132                           & 0.917                           & 33.958          \\ \midrule
\multicolumn{2}{l}{SR-SPECNet}              & 0.168                           & 0.909                           & 30.683                           & 0.092                           & 0.946                           & 33.446                           & 0.077                           & 0.955                           & 34.326           \\ \midrule
\multicolumn{2}{c}{SR-SPECNet+}             & \textbf{0.080} & \textbf{0.950} & \textbf{34.006} & \textbf{0.063} & \textbf{0.962} & \textbf{35.350} & \textbf{0.056} & \textbf{0.965} & \textbf{35.884} \\  \bottomrule
\end{tabular}}
\caption{Performance metrics of deep learning models employing 10 antenna elements for sper-resolution RA heatmap generation. 
}\label{10}
\end{table}

Shifting the focus to the 40-element antenna configuration, as presented in Table \ref{40}, SR-SPECNet+ maintains its exceptional standard. It substantially betters its NMSE and improves SSIM and PSNR values, asserting its robustness across all data volumes. SR-SPECNet retains a strong performance profile, lending weight to the concept that a 1D super-resolution approach effectively facilitates the production of high-quality heatmaps without extensive training data. This evidence underscores our model's proficiency in handling data from varying antenna aperture sizes and confirms the strategic value of the 1D super-resolution methodology, which maximizes training data efficiency and is conducive to high-resolution radar imaging applications.

\begin{table}[]
\resizebox{\linewidth}{!}{
\begin{tabular}{clccccccccc}
\toprule
\multicolumn{2}{c}{\multirow{2}{*}{Models}} & \multicolumn{3}{c}{small}                           & \multicolumn{3}{c}{medium}                          & \multicolumn{3}{c}{large}                           \\ \cmidrule{3-11} 
\multicolumn{2}{c}{}                        & NMSE$\downarrow$ & SSIM$\uparrow$ & PSNR$\uparrow$  & NMSE$\downarrow$ & SSIM$\uparrow$ & PSNR$\uparrow$  & NMSE$\downarrow$ & SSIM$\uparrow$ & PSNR$\uparrow$  \\  \midrule
\multicolumn{2}{c}{2D U-Net}                & 0.233                        &  0.903                        & 34.676                          &  0.173                       & 0.934                         & 36.276                          & 0.124                        & 0.941                         & 36.887        \\ \midrule

\multicolumn{2}{c}{3D U-Net} & 1.255 &  0.813 & 31.649 & 1.320 &  0.816 & 33.106  & 0.292 & 0.946  &  37.378           \\ \midrule

\multicolumn{2}{l}{SR-SPECNet}             & 0.268  &   0.929   & 34.135   & 0.170  & 0.942  & 36.084  & 0.115  &  0.960  & 38.100          \\ \midrule
\multicolumn{2}{c}{SR-SPECNet+} & \textbf{0.144}                        & \textbf{0.951}                         & \textbf{36.924}                          & \textbf{0.122}                        & \textbf{0.959}                         & \textbf{37.916}                          & \textbf{0.093}                        & \textbf{0.965}                        &  \textbf{39.066} \\  \bottomrule
\end{tabular}}
\caption{Performance metrics of deep learning models  employing 40 antenna elements for super-resolution RA heatmap generation. 
}\label{40}
\end{table}

\subsection{Complexity and Scalability}

The 2D and 3D U-Nets, designed for processing low-resolution RA and RAD heatmaps, have fixed numbers of trainable parameters for different size antenna arrays: approximately 31.0M for the 2D U-Net and 51.8M for the 3D U-Net. Their inference times are 6.14 ms and 6.98 ms, respectively, indicating the increased computational demands of higher-dimensional data processing. In contrast, our SR-SPECNet achieves notable efficiency, operating with just 1M parameters and a swift inference time of 3.12 ms. This reduction in time and model size enhances the speed and resource efficiency of our method, ideal for real-time automotive radar applications on embedded CPUs. Meanwhile, the IAA's inference times are 12.3 ms for 10-element vectors and 30.73 ms for 40-element vectors, underscoring its computational challenges.

\begin{figure*}%[h]
\centering
\includegraphics[width=\linewidth]{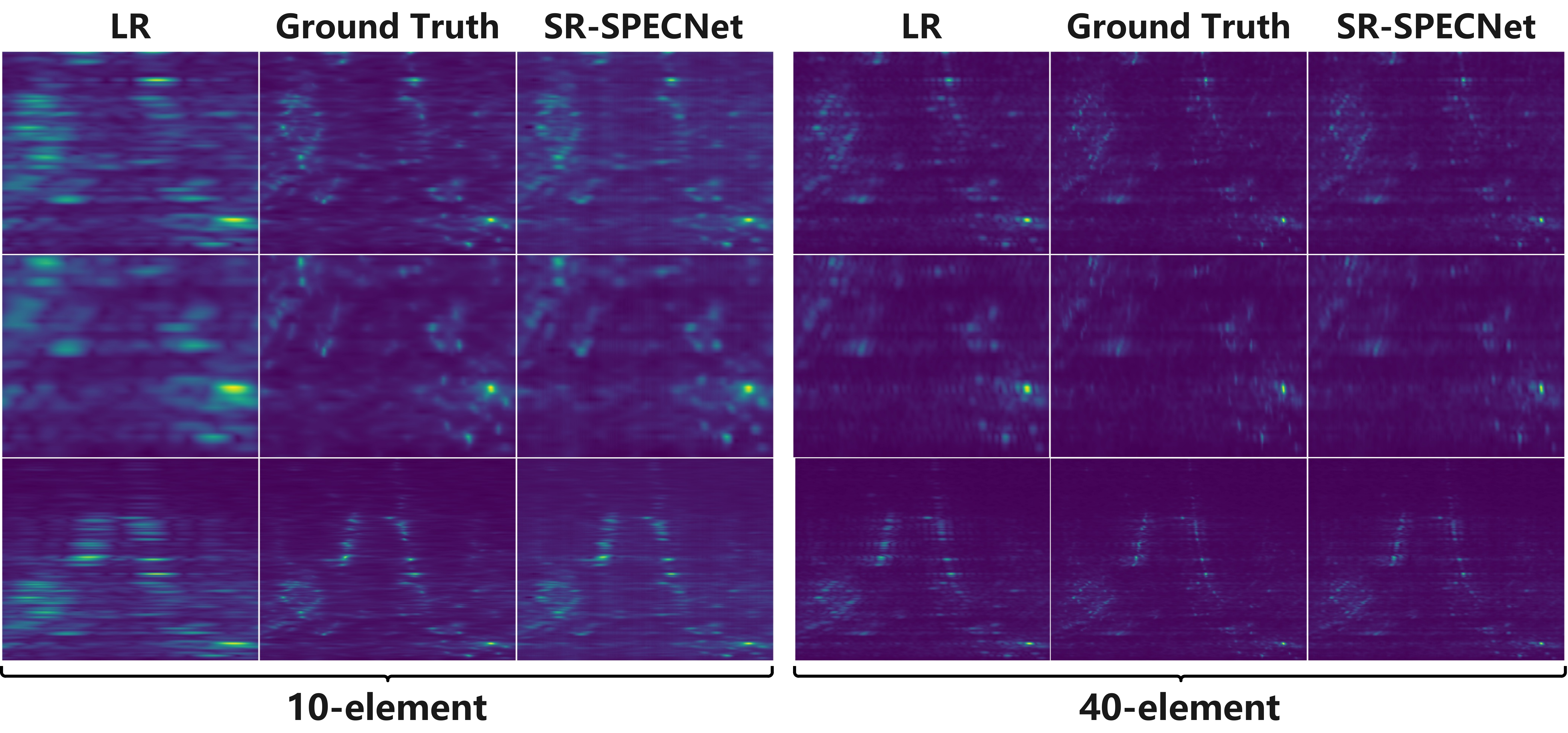}
\caption{\label{scale} Scalability of SR-SPECNet across variable $N_{\text{Range}}$ and $N_{\text{Doppler}}$. The figure contrasts RA heatmap reconstructions for 10 and 40 antenna elements,  respectively.  In the first row, $N_{\text{Range}} = 100$, $N_{\text{Doppler}} = 64$; in the second row, $N_{\text{Range}} = 50$, $N_{\text{Doppler}} = 40$; and in the third row, $N_{\text{Range}} = 200$, $N_{\text{Doppler}} = 40$, showcasing the model's scalability.}
\end{figure*}

Our SR-SPECNet model is designed for adaptability, accepting 1D beam vectors as input, which provides superior scalability across varying $N_{\text{Range}}$ and $N_{\text{Doppler}}$ values. As outlined in Section \ref{sec:experiment:imp}, our training dataset is configured with $N_{\text{Range}} = 100$ and $N_{\text{Doppler}} = 64$. However, as demonstrated in Figure~\ref{scale}, SR-SPECNet effortlessly handles different sizes of these parameters. We generate low-resolution and ground truth heatmaps using FFT and the IAA algorithm, respectively. This remarkable scalability highlights the advantages of our approach, confirming its suitability for diverse and dynamic radar imaging scenarios.

\subsection{Generalizability}
Our training and test datasets feature a rich variety of signals, with each radar frame containing thousands of signals to ensure diversity. To further test the model's generalizability, we evaluated our pre-trained SR-SPECNet on 10,500 radar frames from Radartron \citep{radatron}, which present different $N_{\text{Range}}$ and  $N_{\text{Doppler}}$. Unlike the 2D and 3D UNets, which could not be applied directly due to these variations, our 1D approach was seamlessly implemented, showcasing its superior scalability. Performance metrics include a NMSE of 0.025, SSIM of 0.983, and PSNR of 37.858 for the 10-element configuration, and NMSE of 0.0743, SSIM of 0.970, and PSNR of 42.941 for the 40-element configuration.

\begin{figure*}[h]
\centering
\includegraphics[width=1.0\linewidth]{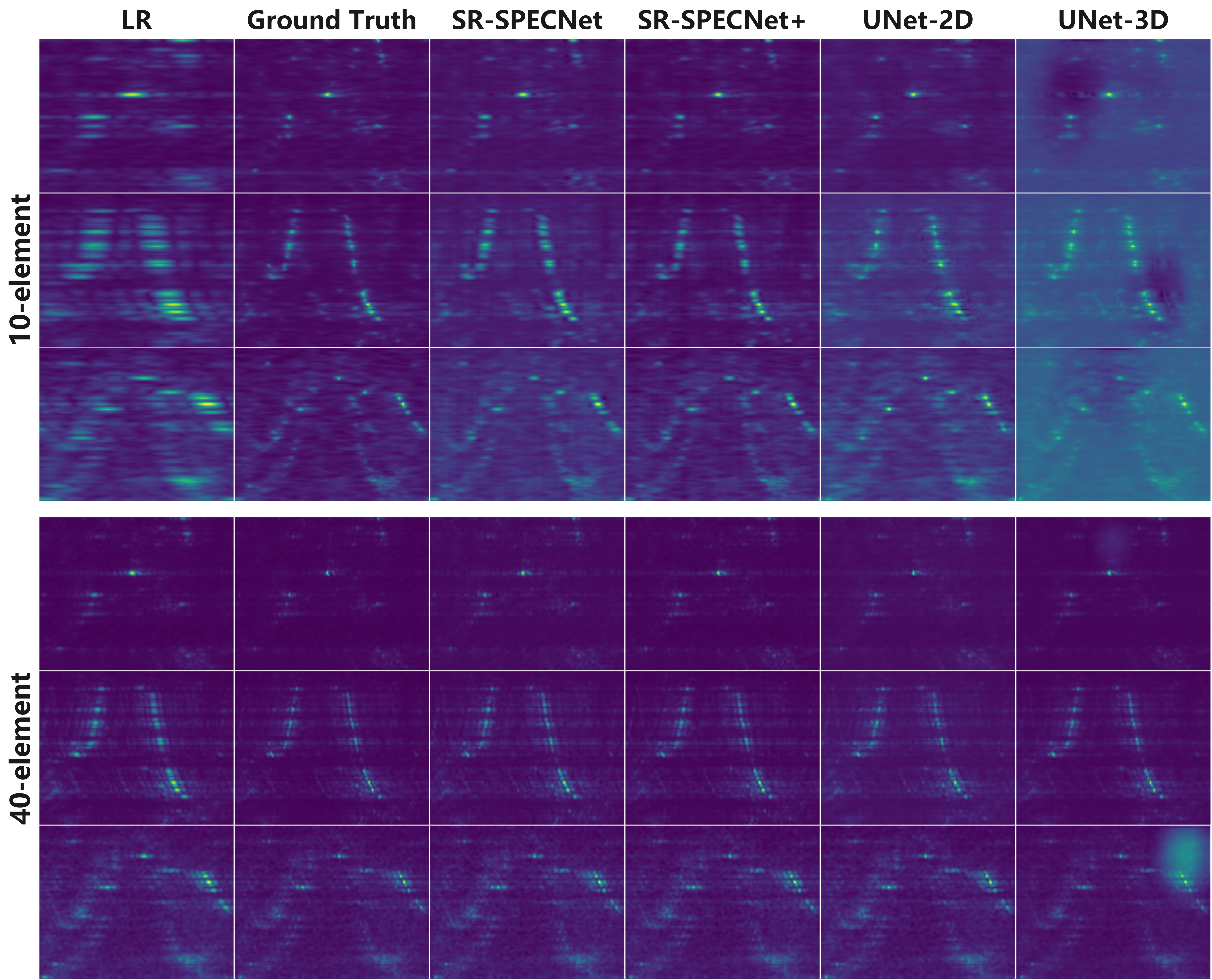}
\caption{\label{results} RA heatmap quality comparison. Heatmaps from the same radar frame are reconstructed by SR-SPECNet, SR-SPECNet+, and baseline 2D U-Net and 3D U-Net models, alongside the ground truth, for both 10-element and 40-element antenna arrays. Each row corresponds to heatmaps generated from the same radar frame data.}
\end{figure*}

\subsection{Visualization of RA maps}
To evaluate the quality of high-resolution RA heatmap by SR-SPECNet and SR-SPECNet+, we visually compare the outputs against those from baseline models, i.e., 2D U-Net and 3D U-Net, as depicted in Fig. \ref{results}. Each model is trained on the `large' dataset to obtain the best performance. Ground truth heatmaps generated with the IAA, have much better resolution than the LR images.  
Further, IAA suppresses sidelobes, yielding much clearer heatmaps. This underscores the benefits of using IAA-derived as ground truth for learning, rather than using FFT-generated heatmaps with a larger antenna aperture, which may not always be practically available due to hardware cost and the complexities involved in MIMO technology implementations.
Fig. \ref{results} demonstrates that  
the RA heatmaps that are respectively produced by SR-SPECNet and  SR-SPECNet+ with a SNR-guided loss function, exhibit a noticeable improvement over the heatmaps generated by 2D and 3D U-Net, aligning with the quantitative metrics presented in Tables \ref{10} and \ref{40}. These results collectively affirm the effectiveness of our proposed 1D methodologies over the 2D and 3D methods.

\section{Conclusions}
\label{sec:conclusion}

%--------------------------------------------------------------------
In this study, we have advanced automotive radar imaging by introducing SR-SPECNet, a novel 1D network leveraging IAA-generated RA heatmaps as ground truth and integrating a unique SNR-guided loss function for super-resolution RA heatmap generation. Our approach, emphasizing a 1D signal processing perspective, has demonstrated superior performance on real radar measurements in terms of automotive radar imaging quality, scalability, and efficiency across varying antenna configurations and dataset sizes. These contributions enhance the fidelity of radar imaging in autonomous vehicles. They also opens avenues for future research, especially with our commitment to sharing our own radar dataset and source code resources with the research community. This work underscores the potential of deep learning-enhanced radar processing in improving navigational safety and robust perception in autonomous vehicles.
\section{Limitation \& Future works}
Our proposed method opens up possibilities for super-resolution imaging across four-dimensional (4D) radar parameters, range, Doppler, azimuth, and elevation. Extending our method to fully support 4D radar imaging represents a key direction for our future research. Currently, our network is optimized for ULA only; thus, adapting our model to accommodate arbitrary array geometries, including various sparse configurations and arrays affected by random antenna failures, is crucial. Additionally, we plan to enhance our network’s performance across diverse dynamic ranges and in scenarios characterized by extremely low SNR. These improvements are vital for advancing the robustness and applicability of our technology in real-world environments.

\bibliographystyle{abbrvnat}
\bibliography{main}
\end{document}